\documentclass[letterpaper]{article} 
\usepackage{aaai24}  
\usepackage{times}  
\usepackage{helvet}  
\usepackage{courier}  
\usepackage[hyphens]{url}  
\usepackage{graphicx} 
\usepackage{natbib}  
\usepackage{caption} 
\usepackage{algorithm}
\usepackage{algorithmic}

\usepackage{amsmath}
\usepackage{amssymb}

\usepackage{makecell,rotating}
\usepackage{multirow}
\usepackage{booktabs}

\usepackage{bbding}
\usepackage{pifont}

\usepackage[table,xcdraw]{xcolor}

\usepackage{newfloat}
\usepackage{listings}
\DeclareCaptionStyle{ruled}{labelfont=normalfont,labelsep=colon,strut=off} 
\floatstyle{ruled}
\newfloat{listing}{tb}{lst}{}
\floatname{listing}{Listing}
\title{Towards Efficient and Effective Text-to-Video Retrieval with Coarse-to-Fine Visual Representation Learning}
\author{
    Kaibin Tian,
    Yanhua Cheng,
    Yi Liu,
    Xinglin Hou,
    Quan Chen,
    Han Li
}
\affiliations{
    Kuaishou Technology \\
    \{tiankaibin, chengyanhua, liuyi24, houxinglin, chenquan06, lihan08\}@kuaishou.com

}

\begin{document}

\maketitle

\begin{abstract}
In recent years, text-to-video retrieval methods based on CLIP have experienced rapid development. The primary direction of evolution is to exploit the much wider gamut of visual and textual cues to achieve alignment. Concretely, those methods with impressive performance often design a heavy fusion block for sentence (words)-video (frames) interaction, regardless of the prohibitive computation complexity. Nevertheless, these approaches are not optimal in terms of feature utilization and retrieval efficiency. 
To address this issue, we adopt multi-granularity visual feature learning, ensuring the model's comprehensiveness in capturing visual content features spanning from abstract to detailed levels during the training phase. To better leverage the multi-granularity features, we devise a two-stage retrieval architecture in the retrieval phase. This solution ingeniously balances the coarse and fine granularity of retrieval content. Moreover, it also strikes a harmonious equilibrium between retrieval effectiveness and efficiency.
Specifically, in training phase, we design a parameter-free text-gated interaction block (TIB) for fine-grained video representation learning and embed an extra Pearson Constraint to optimize cross-modal representation learning. In retrieval phase, we use coarse-grained video representations for fast recall of top-k candidates, which are then reranked by fine-grained video representations.
Extensive experiments on four benchmarks demonstrate the efficiency and effectiveness. Notably, our method achieves comparable performance with the current state-of-the-art methods while being nearly 50 times faster. 
\end{abstract}

\section{Introduction}
\begin{figure}[t]
\begin{center}
\includegraphics[width=1.0\linewidth]{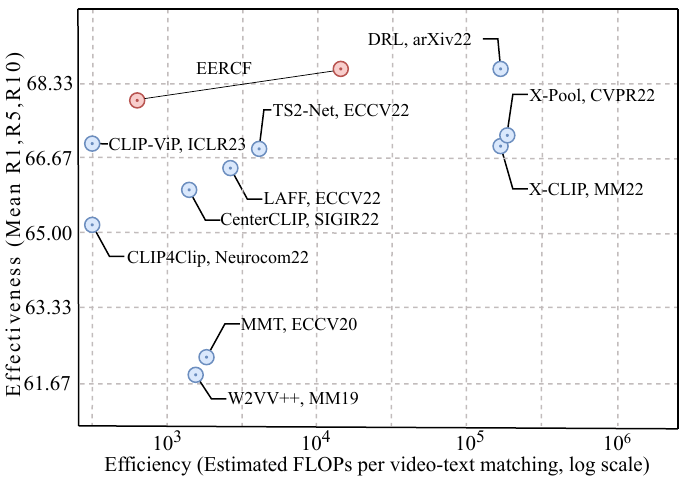}
\end{center}
\caption{Effectiveness and efficiency for text-to-video retrieval
models. We evaluate our approach under the settings of MSRVTT-1K-Test and backbone CLIP(ViT-B/32). The current trend of mainstream  is reflected  from the lower left to the upper right corner. Our method achieves the best balance, positioned at the upper left corner.}
\label{fig:introduction}
\end{figure}

With the explosive growth of videos uploaded online every day from platforms like TikTok, Kwai, YouTube, and Netflix, text-to-video retrieval is a crucial and fundamental task for multi-modal representation learning~\cite{clip2video,bridgeformer,xpool,clip4clip,xclip}. 
Recently, the pre-trained text-image matching models (CLIP~\cite{clip}) from a large scale of web-collected image-text pairs show the great success on diverse vision-language downstream tasks~\cite{nichol2021glide,ramesh2022hierarchical,mokady2021clipcap,hu2022scaling,li2022blip,conde2021clip}. In light of the well-learned visual features, a preliminary study is conducted by CLIP4Clip~\cite{clip4clip}, which transfers the pre-trained CLIP to video-language domain with simple MeanPooling and achieves a certain level of promotion. 
A problem arises where, unlike relatively less visual information in text-image matching, aggregating the entire video as a representation in text-to-video retrieval can lead to over-abstraction and be misleading. As one sentence generally describes video sub-regions of interest.
Therefore, a natural idea is to consider how to align text and video representations at a finer granularity.
The development of CLIP variants can be divided into two main categories to address the above problem. One category focuses on designing a heavy fusion block to strengthen the interaction between visual (video, frames) and text (sentence, words) cues for better alignment of the two modalities~\cite{liu2021hit,xclip}.
The other one optimizes text-driven video representations by keeping multi-grained features including video-level and frame-level for brute-force search~\cite{xpool,wang2022disentangled}. 

Despite the big promotion, such mechanisms bring a non-negligible increase in computational cost for real applications, as shown in Fig.\ref{fig:introduction}.
The computational cost here refers to text-video similarity calculation, when a video (text) is represented by one or more vectors. This determines the efficiency of retrieval in practical applications. 
Additionally, we also believe that excessively fine-grained calculations may amplify the noise in local parts of the video, resulting in reduced retrieval effectiveness.

To make a better trade-off between effectiveness and efficiency for text-to-video retrieval, this paper proposes a novel method, namely EERCF, 
towards coarse-to-fine adaptive visual representations learning following a recall-then-rerank pipeline.
Towards coarse-to-fine adaptive visual representations, we adopt the basic framework of CLIP4Clip~\cite{clip4clip} and design a parameter-free text-gated interaction block (TIB) for fine-grained video representation learning, which further refines the granularity from frame to patch compared with other methods. 
Concretely, TIB can adjust the weights of different frame-level features or patch-level features given a text and aggregate them into a video representation, making the matching between text and video more accurate.
In the learning process,  we employ a joint inter- and intra-feature supervision loss following DRL~\cite{wang2022disentangled}, satisfying both cross-modal feature matching and redundancy reduction across feature channels. Inspired by the success of Pearson Constraint in knowledge distillation~\cite{huang2022knowledge}, we adapt it to reduce redundancy across feature channels.

Taking efficiency into consideration, a two-stage retrieval strategy is adopted when using coarse-to-fine visual representations in practice. There are three levels of video representations, including text-agnostic features without text interaction and text-driven aggregation of frame-level and patch-level features generated from TIB module. The text-agnostic video representations are used for fast recall of top-k candidates, which are then reranked by another two fine-grained video representations. Besides, we found that using coarse-grained features in the recall stage can avoid the noise introduced by overly fine-grained features which typically pay more attention to visual details, thereby improving retrieval performance.




Therefore, our approach is towards \textbf{E}fficient and \textbf{E}ffective  text-to-video \textbf{R}etrievl  with \textbf{C}oarse-to-\textbf{F}ine visual representation learning and is coined \textbf{EERCF}. We summarize our main contributions as follows:

1) We introduce a text-gated interaction block without extra learning parameters for multi-grained adaptive representation learning, whilst introducing a combination of inter-feature contrastive loss and intra-feature Pearson Constraint for optimizing feature learning.

2) We propose a two-stage text-to-video retrieval strategy that strikes the optimal balance between effectiveness and efficiency, facilitating the practical implementation.

3) Our method EERCF achieves comparable performance with the current state-of-the-art(SOTA) methods while our FLOPs for cross-modal similarity matching in MSRVTT-1K-Test, MSRVTT-3K-Test, VATEX, and ActivityNet are 14, 39, 20 and 126 times less than the SOTA.




\begin{figure*}[thb!]
\begin{center}
\includegraphics[width=1.0\linewidth]{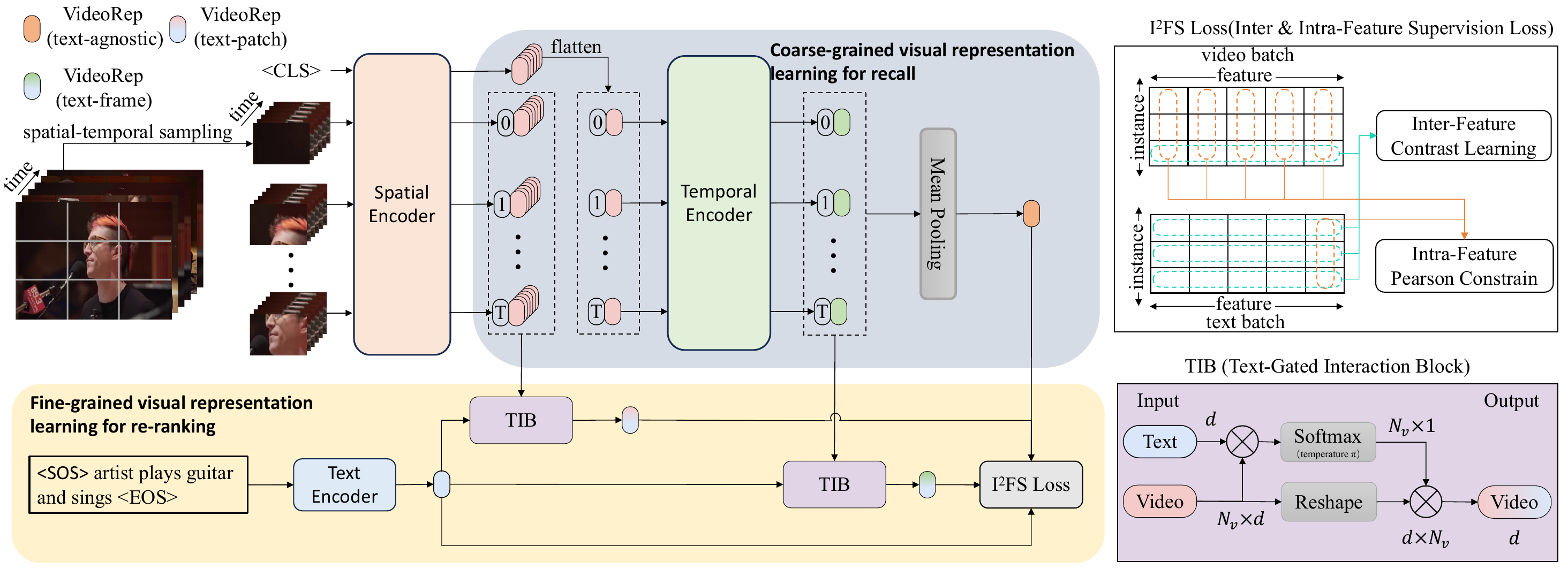}
\end{center}
\caption{Overview of the proposed EERCF framework. EERCF mainly consists of two parts: 1) Coarse-grained and fine-grained visual representations obtained from the TIB module for the recall-reranking pipeline. 2) Inter- and intra-feature supervision loss for optimizing representation learning.  Best viewed in color.}
\label{fig:framework}
\end{figure*}


\section{Related Work}

\subsection{Cross-model Representation Optimization}
The pioneering work, CLIP~\cite{clip}, collects 400M public image-text pairs on the internet and demonstrates the great power of visual-linguistic representation on various downstream tasks, including text-to-image generation~\cite{nichol2021glide,ramesh2022hierarchical}, image caption~\cite{mokady2021clipcap,hu2022scaling} and vision understanding~\cite{li2022blip,conde2021clip}. Benefiting from the pre-trained CLIP model, CLIP4Clip~\cite{clip4clip} adapts it to video-text retrieval with MeanPooling for aggregating video features while still outperforming models pre-trained on video data~\cite{bain2021frozen,xu2021videoclip,xue2022advancing}. CLIP and its variants all employ contrastive learning for cross-modal feature alignment. Recently, DRL~\cite{wang2022disentangled} demonstrates the optimization of correlation reduction for video-text retrieval via a regularization trick. Therefore, We embed the extra Pearson Constraint as a trick to reduce correlation among different cross-modal feature channels, resulting in optimized video and text features. 

\subsection{Boosting Video Representation from Text Interaction}
Note that some CLIP variants, such as CLIP4Clip, TS2-Net, and CLIP-VIP~\cite{clip4clip,clipvip}, for video-text retrieval utilize two independent encoders to represent the two modalities for efficient deployment. For lack of text interaction, the video representation can be over-abstract and misleading to match the common sub-cues depicted by multiple corresponding texts. To address this gap, recent works, such as X-CLIP, TS2-Net, X-Pool and DRL~\cite{xclip,ts2net,xpool,wang2022disentangled}, pay attention to adaptive video features with different text interaction mechanisms. X-CLIP and TS2-Net~\cite{xclip} design a heavy interaction block with multi-grained attention for joint video and text representation learning. X-Pool and DRL~\cite{xpool,wang2022disentangled} improve performance by redesigning the interaction block with a few learning parameters while computing similarity measures between multiple frame features and the query sentence feature for exhaustive search. Our approach goes one step further, developing a text-gated interaction block without extra learning parameters and generating more fine-grained levels of video features, including a text-agnostic version and text-driven aggregation of patch-level and frame-level versions. 
The introduction of more fine-grained features has a worse impact on efficiency. Therefore, it is inevitable to adopt an efficient two-stage strategy.


\subsection{Re-ranking for Cross-Modal Retrieval}

Re-ranking in cross-modal retrieval is not uncommon. For instance, in text-image retrieval tasks, \cite{wei2020boosting} proposes in the embedding space MVSE++ to rerank the recalled candidates using k nearest neighbors. This is a typical unsupervised re-ranking method, as the re-ranking process is independent of the text modality. However, in the field of text-video retrieval, there are not many methods for re-ranking. Our method provides a paradigm for implementing re-ranking. Specifically,
the text-agnostic coarse-grained video features are used for fast recall, and the text-driven fine-grained features benefit to high-performance re-ranking, thus making a better trade-off between efficiency and effectiveness. As the re-ranking process is related to the text modality, we refer to it as text-supervised re-ranking.



\section{Task Formulation}
This paper mainly focuses on the task of text-to-video retrieval ($t2v$), while also taking into consideration the task of video-to-text retrieval ($v2t$).
In $t2v$, the objective is to rank all videos from a video gallery set $\mathcal{V}$ given a query text $t$ based on a similarity score function $s(t,v)$. 
The $v2t$ task is the reverse of $t2v$.
In both tasks, the gallery set is provided ahead of time for retrieval.
Specifically, a video $v$ consists of $T$ sequential frames $\{f_{1},f_{2},...,f_{T} | f_{i} \in \mathbb{R}^{^{\mathit{H\times W\times C}}} \}$, where each frame is divided into $N$ patches $ \{f_{i}^{1},f_{i}^{2},...,f_{i}^{N} | f_{i}^{n} \in \mathbb{R}^{^{\mathit{P\times P \times C}}}\}$ with $\mathit{P \times P}$ size. A text $t$ is defined as a sequence of tokenized words.

\section{Methodology}

In this section, we first introduce the overall architecture of coarse-to-fine
visual representation learning in "Overall Architecture". 
Then we provide a detailed exposition of the parameter-free text-gated interaction block(TIB). 
Next, we explain the joint inter- and intra-feature supervision loss function to optimize cross-modal feature learning.
Finally, we present a two-stage retrieval strategy utilizing well-extracted multi-grained features.


\subsection{Overall Architecture} \label{ss:overarchitecture}

Fig.\ref{fig:framework} illustrates the overall architecture of the proposed EERCF framework. 
There are three levels of video representations that need to be learned.
The first one is a text-agnostic video representation. We follow the same setting of CLIP4Clip~\cite{clip4clip}, which utilizes a spatial encoder (SE) with 12 transformer layers initialized by the public CLIP checkpoints, followed by a temporal encoder (TE) with 4 transformer layers to model temporal relationship among sequential frames and a MeanPooling layer (MP) to aggregate all frame-level features into a text-agnostic feature vector. The procedure above can be formulated as follows:

\begin{small}
\begin{align}
\varphi([f_{i}^{0};f_{i}^{1};...;f_{i}^{N}]) & = {\mathit{\rm{\textbf{SE}}}}([f_{i}^{0};f_{i}^{1};...;f_{i}^{N}]+E_{\mathit{spos}}) \\
\phi ([f_{1};f_{2};...;f_{T}]) & = {\mathit{\rm{\textbf{TE}}}}([\varphi(f_{1}^{0});...;\varphi(f_{T}^{0})]+E_{\mathit{tpos}}) \\
\boldsymbol{v}_{\mathit{L_1}} & = \mathit{\rm{\textbf{MP}}}(\phi(f_1),\phi(f_2),...,\phi(f_T))
\end{align}
\end{small}

\noindent where $\varphi(f_{i}^{n}) \in \mathbb{R}^{D}$ denotes the $\mathit{n_{th}}$ patch feature of the $\mathit{i_{th}}$ frame except that $\varphi(f_{i}^{0})$ is the [CLS] token prediction to encode all patch features within $\mathit{i_{th}}$ frame, $\phi (f_{i}) \in \mathbb{R}^{D}$ denotes the $\mathit{i_{th}}$ frame feature, and $\boldsymbol{v}_{\mathit{L_1}} \in \mathbb{R}^{D}$ is the resulted text-agnostic video feature. 
We add spatial position embeddings $E_{\mathit{spos}} \in \mathbb{R}^{D}$ to all the patch embeddings. 
Then all the outputs $\varphi(f_{i}^{0})$ along with temporal position embeddings $E_{\mathit{tpos}} \in \mathbb{R}^{D}$ are loaded into TE to learn temporal relations among frames of a video. 
To further capture particularly visual cues conditioned on the text, another two video features are extracted based on a text-gated interaction block (TIB):

\begin{small}
\begin{align}
\boldsymbol{v}_{\mathit{L_2}} & = \mathit{\rm{\textbf{TIB}}}(\phi(f_1),\phi(f_2),...,\phi(f_T),\theta(t)) \\
\boldsymbol{v}_{\mathit{L_3}} & = \mathit{\rm{\textbf{TIB}}}(\varphi(f_{1}^{1}),...,\varphi(f_{1}^{N}),...,\varphi(f_{T}^{1}),...,\varphi(f_{T}^{N}),\theta(t))
\end{align}
\end{small}

\noindent where $\theta(t) \in \mathbb{R}^{D} $ denotes the sentence feature of the query text, which drives the aggregation of all the frame features to capture related spatiotemporal visual cues, resulting in text-frame interaction feature $\boldsymbol{v}_{\mathit{L_2}} \in \mathbb{R}^{D}$, as well as the aggregation of all the patch features to keep more fine-grained and aligned visual cues, building text-patch interaction feature $\boldsymbol{v}_{\mathit{L_3}} \in \mathbb{R}^{D}$. 
Next, we will provide a detailed introduction to learning finer-grained video representations using the TIB module.


\subsection{Text-Gated Interaction Block} \label{ss:gib}
The TIB module is a parameter-free text-gated interaction block to align the fine-grained video features with the query sentence feature, as shown in Fig.\ref{fig:framework}. It is a simple attention mechanism based on the Softmax function with a temperature coefficient. Then the text-driven video features can be rewritten as follows:

\begin{small}
\begin{align}
\boldsymbol{v}_{\mathit{L_2}} & = \sum_{i=1}^{T} Softmax(\phi(f_i)^\top\theta (t)/\pi)\phi(f_i) \\
\boldsymbol{v}_{\mathit{L_3}} & = \sum_{i=1}^{T}\sum_{m=1}^{N} Softmax(\varphi(f_i^m)^\top\theta (t)/\pi)\varphi(f_i^m) 
\end{align}
\end{small}

\noindent where $\pi$ is the temperature, which decides how much visual cues will be kept for video-level feature aggregation based on the softmax similarity from a text to all frames or patches. A small value of $\pi$ only emphasizes those most relevant visual cues, while a large value pays attention to much more visual cues. Considering efficiency, we do not introduce any learning parameters in TIB. 

\subsection{Inter- and Intra-Feature Supervision Loss} \label{ss:i2f}
We train EERCF using mini-batch iterations with each batch of $B$ video-text pairs $\{(v_b,t_b)\}_{b=1}^{B}$. In each pair, the text $t_b$ is a corresponding description of the video $v_b$. For each modality, we extract a feature matrix $\boldsymbol{F} \in \mathbb{R}^{B \times D}$, where a row vector $\boldsymbol{F}_{b,:}$ denotes the feature representation of instance $b$, and a column vector $\boldsymbol{F}_{:,d}$ denotes all instances' feature values at channel $d$. The two types of vectors are separately used for inter-feature alignment and intra-feature correlation reduction between video and text via the following loss functions.

\noindent\textbf{Contrastive Loss for Inter-Feature Supervision.} Contrastive learning is popularly applied in multi-modal learning tasks~\cite{clip,clip4clip}. Similarly, we employ the infoNCE loss by considering video-text matching pairs as positives and other non-matching pairs in the batch as negatives. Specifically, we jointly optimize the symmetric video-to-text and text-to-video losses:

\begin{small}
\begin{align}
\mathcal{L}_{inter}^{t2v} & = \frac{1}{B}\sum_{b_1=1}^{B} InfoNCE(\boldsymbol{F}_{b_1,:}^{(v)},\boldsymbol{F}^{(t)})\\
\mathcal{L}_{inter}^{v2t} & = \frac{1}{B}\sum_{b_1=1}^{B} InfoNCE(\boldsymbol{F}_{b_1,:}^{(t)},\boldsymbol{F}^{(v)})\\
\mathcal{L}_{\mathit{inter}} & = \mathcal{L}_{inter}^{v2t} + \mathcal{L}_{inter}^{t2v}
\end{align}
\end{small}


\noindent\textbf{Pearson Constraint for Intra-Feature Supervision.} Pearson Constraint is successfully exploited for knowledge distillation in research~\cite{huang2022knowledge}. In this paper, we transfer the idea of correlation reduction among intra-features as a trick to optimize feature learning. Concretely, Pearson Constraint is defined as a distance measure function:

\begin{small}
\begin{align}
d_p(\boldsymbol{F}_{:,d_1}^{(v)},\boldsymbol{F}_{:,d_2}^{(t)})=1-\rho_{p}(\boldsymbol{F}_{:,d_1}^{(v)},\boldsymbol{F}_{:,d_2}^{(t)})
\end{align}
\end{small}

\noindent where $\rho_{p}(\boldsymbol{F}_{:,d_1}^{(v)},\boldsymbol{F}_{:,d_2}^{(t)})$ is the Pearson coefficient\footnote{https://en.wikipedia.org/wiki/Pearson\_correlation\_coefficient} among normalized intra-feature channels between video and text. Finally, we define the intra-feature loss function as follows:




\begin{algorithm}[t]
\caption{Recall and Re-ranking during Retrieval}
\label{alg:algorithm}
\textbf{Input}: video gallery set: $\mathcal{V}=\{ v_i \}_{i=1}^{K}$, a query text: $t$\\
\textbf{Output}: the most matching video
\begin{algorithmic}[1] 
\STATE Encode $t$ as a text feature $\theta(t)$\;
\STATE \# Reacll Stage
\FOR{$i\leftarrow 1,K$}
    \STATE Encode $v_i$ as a video-level video feature $\boldsymbol{v}_{\mathit{L_1},i}$\;
    \STATE Compute the similarity score between  $\theta(t)$ and $\boldsymbol{v}_{\mathit{L_1},i}$\;
\ENDFOR
\STATE Select top $k$ highest score videos as the candidate set.\;
\STATE \# Re-ranking Stage
\FOR{$i\leftarrow 1,k$}
    \STATE Encode $v_i$ as a frame-level video feature $\boldsymbol{v}_{\mathit{L_2},i}$\;
    \STATE Encode $v_i$ as a patch-level video feature $\boldsymbol{v}_{\mathit{L_3},i}$\;
    \STATE Compute the similarity score between  $\theta(t)$ and weighted sum ($\boldsymbol{v}_{\mathit{L_1},i},\boldsymbol{v}_{\mathit{L_2},i},\boldsymbol{v}_{\mathit{L_3},i}$)\;
\ENDFOR
\STATE Select the highest score video as the most matching video\;
\end{algorithmic}
\end{algorithm}

\begin{small}
\begin{equation}
\begin{split}
\mathcal{L}_{\mathit{intra}} & = \sum_{d=1}^{D}{\left \| d_p(\boldsymbol{F}_{:,d}^{(v)},\boldsymbol{F}_{:,d}^{(t)}) \right \|}^2 \\
& + \alpha\sum_{d_1=1}^{D}\sum_{d_2 \neq d_1}{\left \| 1 - d_p(\boldsymbol{F}_{:,d_1}^{(v)},\boldsymbol{F}_{:,d_2}^{(t)}) \right \|}^2
\end{split}
\end{equation}
\end{small}

\noindent where ${\left \|  ... \right \|}^2$ denotes L2-norm regularization and $\alpha$ controls the magnitude of correlation reduction term. $\mathcal{L}_{\mathit{intra}}$ can benefit the model by learning particular and orthogonal cues among each feature channel. Pearson Constraint achieves a relaxed
correlation reduction by allowing each feature channel to
have the strongest correlation with itself, without needing
to be completely independent of other channels. It benefits
the model by learning compact video features.


\noindent\textbf{Total Loss Function.} As a result, the overall training loss $\mathcal{L}_{all}$ can be composed of the inter-feature and intra-feature supervision losses, $\mathit{i.e.}$,

\begin{small}
\begin{equation}
\mathcal{L}_{all} = \sum_{\boldsymbol{v} \in \{\boldsymbol{v}_{\mathit{L_1}}, \boldsymbol{v}_{\mathit{L_2}}, \boldsymbol{v}_{\mathit{L_3}}\}}\lambda_{\boldsymbol{v}}(\mathcal{L}_{\mathit{inter}} + \beta\mathcal{L}_{\mathit{intra}})
\end{equation}
\end{small}

\noindent where $\beta$ weights the loss importance between $\mathcal{L}_{\mathit{inter}}$ and $\mathcal{L}_{\mathit{intra}}$, and $\lambda_{\boldsymbol{v}}$ balances the contribution of each level of video feature learning.


\begin{table*}[t]

\centering
\resizebox{0.9\linewidth}{!}{
\begin{tabular}{@{}lllllllllll@{}}
\toprule
\multicolumn{1}{l}{\multirow{2}{*}{Model}}  & \multicolumn{4}{c}{$t2v$ Retrieval} & \multicolumn{4}{c}{$v2t$ Retrieval} & \multicolumn{1}{l}{\multirow{2}{*}{\begin{tabular}[c]{@{}c@{}}FLOPs \\(k=1000)\end{tabular}}} \\ \cline{2-9}
\multicolumn{1}{c}{} & R@1 & R@5 & R@10 & Mean & R@1 & R@5 & R@10 & Mean & \multicolumn{1}{c}{} \\ 
\midrule
\midrule
\multicolumn{10}{c}{\textit{Backbone model: ViT-B/32}} \\
\hline
TeachText~\cite{teachtext} & 29.6 & 61.6 & 74.2 & 55.1 & 32.1 & 62.7 & 75.0 & 56.6 & 0.8k  \\
SEA~\cite{sea} & 37.2 & 67.1 & 78.3 & 60.9 & - & - & - & - &  10.2k \\
W2VV++~\cite{w2vv++} & 39.4 & 68.1 & 78.1 & 61.9 & - & - & - & - &  2.0k \\
BridgeFormer~\cite{bridgeformer} & 44.9 & 71.9 & 80.3 & 65.7 & - & - & - & - & 0.5k \\
LAFF~\cite{laff} & 45.8 & 71.5 & 82.0 & 66.4 & - & - & - & - &  4.1k \\
CLIP4Clip$\dagger$ & 42.8 & 71.6 & 81.1 & 65.2 & 41.4 & 70.6 & 80.5 & 64.2 & \textbf{0.5k} \\
CenterCLIP~\cite{centerclip} & 44.0 & 70.7 & 81.4 & 65.4 & 42.9 & 71.4 & 81.7 & 65.3 & 1.5k \\
X-CLIP$\dagger$ & 46.3 & 72.1 & 81.8 & 66.7 & \textbf{45.9} & 72.8  & 81.2 & 66.7 &  220.9k \\
TS2-Net$\dagger$ & 46.7 & 72.6 & 81.2 & 66.8 & 43.6 & 71.1 & 82.7  & 65.8 &  6.1k \\
CLIP-VIP~\cite{clipvip} & 46.5 & 72.1 & 82.5   & 67.0 & 40.6 & 70.4 & 79.3 & 63.4 &  \textbf{0.5k} \\
X-Pool~\cite{xpool} & 46.9   & 72.8   & 82.2 & 67.3   & 44.4 & 73.3  & \textbf{84.0} & 67.2  &  275.0k \\
DRL~\cite{wang2022disentangled} & 47.4 & \textbf{74.6} &  83.8 &  \textbf{68.6} &  45.3 &  73.9 &  83.3 &  67.5 &  220.4k \\

EERCF (ours) & \textbf{47.8}  &  74.1 &  \textbf{84.1} &  \textbf{68.6} & 44.7  & \textbf{74.2} & 83.9  & \textbf{67.6}&  16.0k \\
\midrule
\midrule
\multicolumn{10}{c}{\textit{Backbone model: ViT-B/16}} \\
\hline
CLIP4Clip$\dagger$  &  46.4 & 72.1 &  82.0  & 66.8  &  45.4  &  73.4  &  82.4  &  67.1  &  \textbf{0.5k} \\
X-CLIP$\dagger$     &  49.3 & 75.8  &  84.8  &  70.0  &  48.9  &  76.8  & 84.5  &  70.1  & 220.9k \\
DRL~\cite{wang2022disentangled}                 &  50.2 & 76.5  & 84.7  & 70.5  &  48.9 & 76.3 &  85.4 & 70.2 &  220.4k \\
DRL*~\cite{wang2022disentangled}                &  53.3 & \textbf{80.3} & \textbf{87.6} &  \textbf{73.7} &  \textbf{56.2} &  \textbf{79.9}  & \textbf{87.4} & \textbf{74.5} & 220.4k \\
EERCF (ours)        &  49.9 & 76.5 & 84.2 & 70.2  & 47.8 &  75.3 & 84.2 & 69.1 & 0.8k \\ 
EERCF* (ours)       &  \textbf{54.1} &  78.8 & 86.9 &  73.2 &  55.0 & 77.8 &  85.7 & 72.8 & 0.8k \\ 

\bottomrule
\end{tabular}
}

\caption{Comparison of retrieval efficiency and effectiveness on the MSRVTT-1K-Test. The best results are shown in \textbf{bold} and the results unavailable are left blank. Methods marked with $\dagger$ are reproduced in this paper with the same experimental settings for fair comparison. * denotes we add the DSL or Q-Norm trick to achieve the best performance in comparison. }

\label{tab:msrvtt}
\end{table*}

%




\subsection{Two-stage Strategy in Retrieval} \label{ss:overarchitecturetwostage}

To balance the efficiency and effectiveness for text-to-video retrieval, we utilize $\boldsymbol{v}_{\mathit{L_1}}$ for fast recall of top-k condidates and then rerank them via  $\boldsymbol{v}_{\mathit{L_2}}$ and $\boldsymbol{v}_{\mathit{L_3}}$. Note that all video and text features are $\mathit{L_2}$ normalized for similarity measure. 
It is worth mentioning that we consider using a two-stage method, partly due to its efficiency improvement, and partly due to the fact that overly fine-grained features may lead to an excessive focus on local noise. 
The two-stage approach strikes a good balance between overly abstract and overly detailed video representations.


\section{Experiments}

We perform experiments on the commonly used benchmark of MSR-VTT~\cite{msrvtt}, VATEX~\cite{vatex}, MSVD~\cite{msvd}, and ActivityNet~\cite{activitynet}. These datasets vary in video duration, content, and text annotations, providing a comprehensive evaluation of different methods. Details are listed in the supplementary material. 
Following existing literature~\cite{teachtext,sea,w2vv++,bridgeformer,laff,clip4clip,centerclip,xpool,xclip,clipvip,ts2net}, we report Recall@1 (R@1), Recall@5 (R@5), Recall@10 (R@10), and mean result of them (Mean) for comparison.
We use FLOPs to evaluate efficiency for text-video similarity calculation, which is calculated by THOP\footnote{https://github.com/Lyken17/pytorch-OpCounter}.
It should be noted that due to the relatively small size of the MSVD\cite{msvd}, the results are in the supplementary material. 







\begin{table*}[t]

\centering
\resizebox{0.9\linewidth}{!}{
\begin{tabular}{@{}l|llll|llll|llll@{}}
\toprule
\multirow{2}{*}{Model} & \multicolumn{4}{c|}{MSRVTT-3K-Test} & \multicolumn{4}{c|}{VATEX} & \multicolumn{4}{c}{ActivityNet} \\ 
& \multicolumn{1}{c}{R@1} & \multicolumn{1}{c}{R@5} & \multicolumn{1}{c}{Mean} & \multicolumn{1}{c|}{FLOPs} & \multicolumn{1}{c}{R@1} & \multicolumn{1}{c}{R@5} & \multicolumn{1}{c}{Mean} & \multicolumn{1}{c|}{FLOPs} & \multicolumn{1}{c}{R@1} & \multicolumn{1}{c}{R@5} & \multicolumn{1}{c}{Mean} & \multicolumn{1}{c}{FLOPs}\\ 
\midrule
TeachText &15.0 & 38.5 & 35.1 & 0.8k & 53.2 & 87.4 & 78.0  & 0.8k  & 23.5 & 57.2 & 58.9 & 0.8k  \\
SEA & 19.9 & 44.3 & 40.2 & 10.2k &  52.4 & 90.2 & 79.5 & 10.2k & - & - & - & - \\
W2VV++ & 23.0 & 49.0 & 44.2 & 2.0k & 55.8 & 91.2 & 81.0 & 2.0k & - & - &  -& - \\
LAFF & 29.1 & 54.9 & 49.9 & 4.1k & 59.1 & \textbf{91.7}  & 82.4 & 4.1k  & - & - & - & - \\
CLIP4Clip$\dagger$ & 29.4 & 54.9 & 50.0 & \textbf{0.5k}  & 61.6 & 91.1 & 82.8 & \textbf{0.5k} &  39.7 & 71.0 & 64.7 & \textbf{0.5k} \\
TS2-Net$\dagger$ & 29.9 & 56.4 & 51.2 & 6.1k & 61.1 & 91.5  & 82.9 & 6.1k & 37.3 & 69.9 & 63.5 & 32.8k\\
X-CLIP$\dagger$ & 31.2 & \textbf{57.4}  & \textbf{52.2} & 220.9k  &  62.1  & 90.8 & 82.7 & 220.9k & \textbf{44.4}  &  \textbf{74.6} & 68.0  & 2175.9k \\
DRL & - & - & - & - & \textbf{63.5} & \textbf{91.7} & \textbf{83.9} & 220.4k & 44.2 & 74.5 & \textbf{68.3} & 2175.4k \\
\midrule
\midrule
EERCF (ours) & \textbf{31.5}  & \textbf{57.4} & \textbf{52.2}  & 5.7k & 62.6  & 91.5  & 83.3 & 10.8k &  43.1 & 74.5 & 67.9 & 17.3k \\
\bottomrule
\end{tabular}
}
\caption{Comparison of text-to-video retrieval efficiency and effectiveness on the MSRVTT-3K-Test, VATEX and ActivityNet. Results on video-to-text retrieval are similar and omitted due to limited space.}
\label{tab:multi_dataset}
\end{table*}

\subsection{Implementation Details}\label{ssec:details}

We perform the experiments on 24 NVIDIA Tesla T4 15GB GPUs using the PyTorch library. Similar to~\cite{clip4clip,xclip}, the spatial encoder and text encoder of EERCF are initialized by the CLIP checkpoints. We train our model via Adam optimizer and decay the learning rate using a cosine schedule strategy. For better finetuning, we set different learning rates for different modules, where the spatial encoder and text encoder are set to 1e-7, owning to CLIP initialization, and other new modules, like the temporal encoder, are set to 1e-4. The max word token length and max frame length are fixed to 32 and 12 for MSR-VTT, MSVD, and VATEX, while the corresponding settings are 64 and 64 for ActivityNet due to longer captions of video-paragraph retrieval. Limited by GPU memory, we set the batch size of MSR-VTT, MSVD, VATEX, and ActivityNet to 240, 240, 360, and 96, respectively.
We train 5 epochs for all datasets. Unless otherwise specified, the hyperparameters mentioned in our equations are empirically set as follows: $\pi=0.1$ and $\pi=0.01$ separately for frame-level and patch-level TIB module, $\{\alpha=0.05\}$ in $\mathcal{L}_{\mathit{intra}}$ loss, $\{\beta=0.001, \lambda_{\boldsymbol{v}_{\mathit{L_1}}}:\lambda_{\boldsymbol{v}_{\mathit{L_2}}}:\lambda_{\boldsymbol{v}_{\mathit{L_3}}}=5:5:1\}$ in the total loss, and top-k=50 for our coarse-to-fine retrieval.

\subsection{Performance Comparison}
On all the datasets, EERCF can achieve a performance close to or even exceed the SOTA methods while maintaining the advantage of retrieval efficiency. We present a detailed comparison of both efficiency and effectiveness on MSRVTT-1K-Test in Tab.\ref{tab:msrvtt}. We conclude the following observation:

$\bullet$ Building on ViT-B/32, EERCF achieves outstanding results of 68.6/67.6 Mean in the $t2v$/$v2t$ task, surpassing stronger competitors such as CLIP-VIP (the baseline version without additional data), X-Pool, X-CLIP, TS2Net and DRL. EERCF achieves a large gain than CLIP4Clip model by +11.7\% (+5.0\%) relative (absolute) improvement on $t2v$ R@1, and +8.0\% (+3.3\%) improvement on $v2t$ R@1.


$\bullet$ Building on ViT-B/16, remarkably, EERCF achieves performance comparable to that of DRL while having a significantly lower computation cost of only 0.8k FLOPs. Specifically, the computation cost of DRL is approximately 275 times higher than that of EERCF. 



To further validate the generalization of EERCF, we evaluate its performance on MSRVTT-3K-Test, VATEX and ActivityNet, as presented in Tab.\ref{tab:multi_dataset}. EERCF achieves considerable performance improvement in an efficient manner, being nearly 39, 20, and 126 times faster than the SOTAs in  MSRVTT-3K-Test, VATEX, and ActivityNet. 
\begin{table}[t]

\setlength{\tabcolsep}{5pt}
\centering
\resizebox{\linewidth}{!}{
\begin{tabular}{l|cccc}
\toprule
Model & R@1 & R@5 & R@10 & Mean  \\ \midrule
EERCF & \textbf{47.8} & \textbf{74.1} & \textbf{84.1} & \textbf{67.8}  \\
-~w/o patch-level feature & 47.2 & 73.2 & 82.2 & 67.5 \\
-~w/o frame-level feature  & 45.0 & 71.4 & 81.4 & 65.9 \\
\midrule
-~w/o all text-driven features & 42.8 & 71.6 & 81.1 & 65.2 \\
\bottomrule
\end{tabular}
}
\caption{Ablation for fine-grained video features in EERCF on MSRVTT-1K-Test dataset.}
\label{tab:intraandter}
\end{table}
\begin{figure}[t]
\begin{center}
\includegraphics[width=0.85\linewidth]{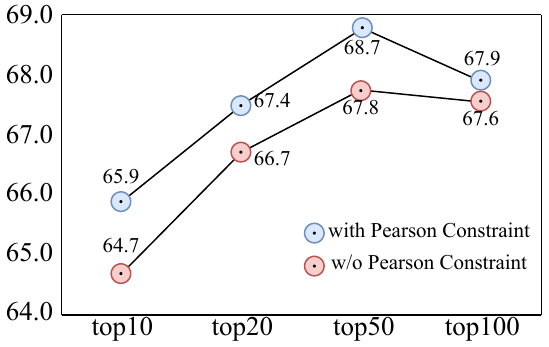}
\end{center}
\caption{Retrieval performance on MSRVTT-1K-Test based on different number of re-ranking candidates k.}
\label{fig:topkfig}
\end{figure}
\subsection{Complexity Analysis}
Normally, we denote the computational complexity as $\mathcal{O}(ND)$ for dot product retrieval, where both video and text are represented by a $D$ dimension vector and the gallery set is with size $N$. When performing more fine-grained retrieval, we set the number of frames as $N_v$, the number of words as $N_t$, the number of patches per frame as $N_p$, and the number of candidate set as $N_r$, which is much smaller than $N$. 
Our complexity, $\mathcal{O}(ND + N_r(1 + N_v + N_p)D)$, is significantly better than DRL's $\mathcal{O}(NN_vN_tD)$ and X-CLIP's $\mathcal{O}(N(N_vN_t+N_v+N_t+1)D)$.

\subsection{Ablation Study}
Since MSR-VTT is more popular and competitive compared to other datasets, we conduct ablation, quantitative and qualitative experiments on it.
In this section, we carefully investigate the proposed EERCF, including the contributions of multi-grained video representations, the effect of Pearson Constraint, the selection of top-k re-ranking hyperparameter.
We further compare the differences between different re-ranking methods and visualize how fine-grained visual representations affect the re-ranking process.
\begin{figure*}[t]
\begin{center}
\includegraphics[width=1.0\linewidth]{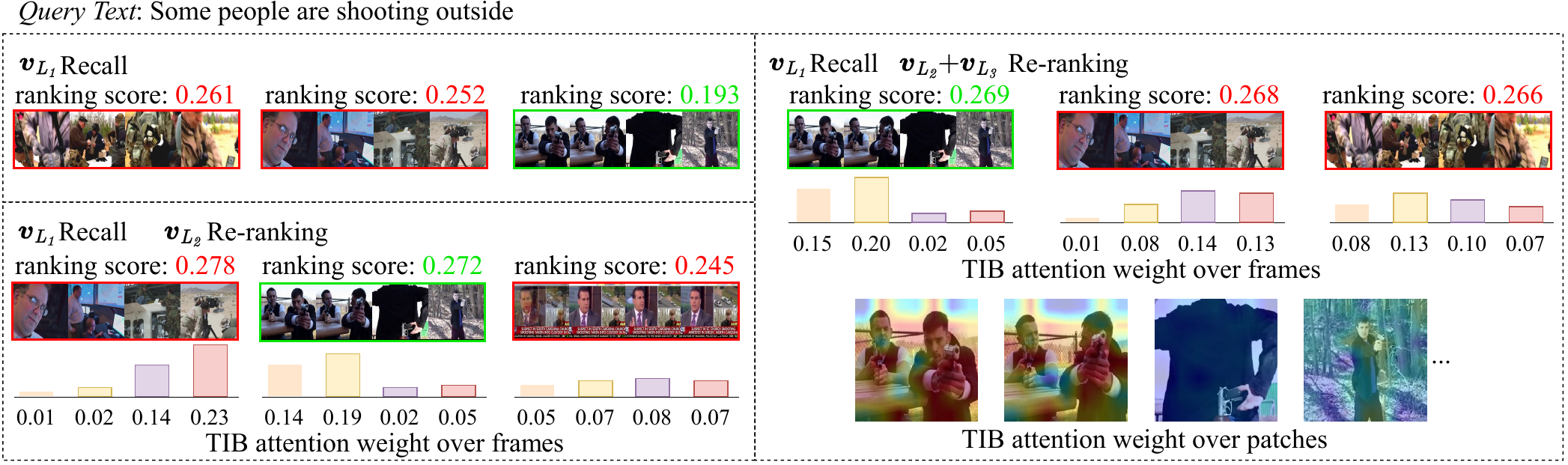}
\end{center}
\caption{Visualization of the coarse-to-fine retrieval process on MSRVTT-1K-Test. Green boxes mean the ground truth video corresponding to the query text, and red boxes denote confused videos. More results are provided in the supplementary material.}
\label{fig:vis}
\end{figure*}

\textbf{Fine-grained text-driven features.}
Ablation results are presented in Tab.\ref{tab:intraandter}. Upon removing either the text-frame interaction feature or the text-patch interaction feature from EERCF, a decline in retrieval performance can be observed, particularly in the removal of the text-frame feature. This decline serves to underscore the effectiveness of both fine-grained features, which adeptly capture distinct levels of visual and temporal cues from the query text. Furthermore, when we exclude all text-driven video features in EERCF and instead opt for the conventional single-stage retrieval paradigm relying solely on the text-agnostic video feature, a significant decrease in retrieval performance becomes evident. The phenomenon demonstrates the limitations inherent in the text-agnostic video feature, which may tend to oversimplify and potentially mislead the matching process for diverse query texts. It also indicates the indispensability of the TIB, responsible for extracting fine-grained adaptive features in the second re-ranking stage.

\begin{table}[t!]

\setlength{\tabcolsep}{5pt}
\centering
\resizebox{\linewidth}{!}{
\begin{tabular}{cllllll}
\toprule
\multicolumn{1}{l}{\multirow{2}{*}{\begin{tabular}[c]{@{}c@{}}Intra-Feature\\ Pearson Constraint\end{tabular}}} & \multicolumn{3}{c}{\begin{tabular}[c]{@{}c@{}} Features \end{tabular}} & \multicolumn{3}{c}{$t2v$ Retrieval}  \\ \cline{2-7}
& $\boldsymbol{v}_{\mathit{L_1}}$ & $\boldsymbol{v}_{\mathit{L_2}}$ & $\boldsymbol{v}_{\mathit{L_3}}$  & R@1 & R@5  & Mean  \\ \hline
No & \checkmark &  &   & 42.8 & 71.6  & 65.2  \\
Yes & \checkmark &  &   & 43.6 & 72.0  & 65.7 \\
No & \checkmark & \checkmark &  & 47.2 & 73.2  & 67.5 \\
Yes & \checkmark & \checkmark &  & 47.3 & 73.2 & 67.8 \\
No & \checkmark & \checkmark & \checkmark & 47.8 & 73.0  & 67.8 \\
Yes & \checkmark & \checkmark & \checkmark  & \textbf{47.8} & \textbf{74.1}  & \textbf{68.7} \\ 
\bottomrule
\end{tabular}
}
\caption{Ablation study of Pearson Constraint in EERCF on MSRVTT-1K-Test dataset.}
\label{tab:ifpc}

\end{table}

\textbf{Top-k re-ranking hyper-parameter.} 
As shown in Fig.\ref{fig:topkfig}, EERCF obtains stable retrieval performance improvement when top-k ranges from 10 to 50. We also observe that there is a slight decrease in retrieval performance when the top-k increases to 100. We believe this is due to the re-ranking stage excessively focusing on visual local details and the noise introduced by increasing the top-k can cause more disturbances to the re-ranking stage. And there is some preliminary evidence that X-CLIP also benefits from two-stage retrieval, as the original X-CLIP essentially re-ranking all videos, whereas our experiments only re-ranking the top-50 videos as shown in Tab.\ref{tab:multi_dataset} and Tab.\ref{rerankingcompare}. Also with Pearson Constraint, we have observed consistent performance improvement across different top-k values in retrieval. 


\textbf{Intra-Feature Pearson Constraint.}
We also demonstrate the effectiveness of the Intra-Feature Pearson Constrain through experiments. As shown in Tab.\ref{tab:ifpc}, the retrieval performance has shown consistent improvement through the use of the Intra-Feature Pearson Constrain from coarse to fine video features. The above-mentioned results have proven the effectiveness of using the Pearson Constrain to enhance the uniqueness of features and reduce redundant information.

\textbf{Different Re-ranking Methods.}
Our efficiency improvement is primarily due to the second stage of re-ranking. So we evaluate our method against previously computationally expensive methods in the second stage. We also compare with a commonly used unsupervised re-ranking method named Video Sim\cite{wei2020boosting}. 
Each video in the candidate set undergoes a voting process based on inter-video similarity, with the resulting vote count determining the final re-ranking order.
To clarify, we perform re-ranking on the same candidate set.
 As shown in Tab. \ref{rerankingcompare}, our method outperforms the alternatives in both performance and efficiency.

\begin{table}[t!]
\setlength{\tabcolsep}{3pt}
\resizebox{\linewidth}{!}{
\begin{tabular}{llllll}
\toprule
\multicolumn{2}{c}{\multirow{2}{*}{\begin{tabular}[c]{@{}c@{}}Reranking Method\end{tabular}}} & \multicolumn{3}{c}{Performance} & \multirow{2}{*}{\begin{tabular}[c]{@{}c@{}}Efficiency\\ (FLOPs)\end{tabular}} \\ \cline{3-5}
\multicolumn{2}{l}{} & R@1 & R@5 & Mean &  \\ \hline
\multirow{3}{*}{\begin{tabular}[c]{@{}c@{}}Text\\ Supervised\end{tabular}} & X-CLIP &  46.8 & 72.8 & 66.9 & 11.5k \\ 
 & DRL & 45.2 & 73.1 & 67.4 &  11.6k \\ 
 & EERCF($\boldsymbol{v}_{\mathit{L_2}}$) & \textbf{47.3} & \textbf{73.2} & \textbf{67.8} & \textbf{0.8k} \\  
\hline
Unsupervised & Video Sim & 42.5 & 68.7 & 63.2 & 1.8k \\
\bottomrule
\end{tabular}
}
\caption{Ablation study of different re-ranking methods on MSRVTT-1K-Test dataset.}
\label{rerankingcompare}
\end{table}

\textbf{Visualization of Coarse-to-Fine Retrieval.}
We illustrate the coarse-to-fine reranking in Fig.\ref{fig:vis}, which shows text-agnostic video features often find confused results, and TIB can gradually dig relevant frame-level or patch-level cues for accurate results. We observe that fine-grained information can lead to more precise matching with query texts, but inevitably introduces some noise. This also highlights the necessity of two-stage retrieval, which strikes a good balance between overly abstract and detailed video representations.

\section{Conclusion}
This paper develops a novel EERCF framework for efficient and effective text-to-video retrieval with coarse-to-fine visual representation learning.
To this end, a parameter-free text-gated interaction block is exploited to fine-grained video representations. At the same time, we use a Pearson coefficient trick to optimize representation learning. Finally, using a coarse-to-fine retrieval strategy, our approach achieves the best trade-off between performance and cost on all popular datasets.
 

\section{Supplementary Material}

\subsection{Details of the Pearson Constraint }

Pearson Constraint is defined as a distance measure function:
\begin{small}
\begin{equation}
\begin{split}
\rho_{p}(\boldsymbol{F}_{:,d_1}^{(v)},\boldsymbol{F}_{:,d_2}^{(t)}) =\frac{Cov(\boldsymbol{F}_{:,d_1}^{(v)},\boldsymbol{F}_{:,d_2}^{(t)})} {Std(\boldsymbol{F}_{:,d_1}^{(v)})Std(\boldsymbol{F}_{:,d_2}^{(t)})} \\
=\frac{\sum_{b=1}^{B}(\boldsymbol{F}_{b,d_1}^{(v)}-\bar{\mu}_{d_1}^{(v)})(\boldsymbol{F}_{b,d_2}^{(t)}-\bar{\mu}_{d_2}^{(t)})}{\sqrt{\sum_{b=1}^{B}(\boldsymbol{F}_{b,d_1}^{(v)}-\bar{\mu}_{d_1}^{(v)})^2(\boldsymbol{F}_{b,d_2}^{(t)}-\bar{\mu}_{d_2}^{(t)})^2}}
\end{split}
\end{equation}
\end{small}

\noindent where $Cov$ is the covariance, $Std$ is the standard derivation, and $\bar{\mu}_{d}$ denotes the mean channel vector of a batch. Note that the Pearson coefficient has a good scale-invariant property for similarity measure, namely

\begin{small}
\begin{equation}
\begin{split}
\rho_{p}(\boldsymbol{F}_{:,d_1}^{(v)},\boldsymbol{F}_{:,d_2}^{(t)}) =\rho_{p}(m_1\boldsymbol{F}_{:,d_1}^{(v)}+n_1,m_2\boldsymbol{F}_{:,d_2}^{(t)}+n_2)
\end{split}
\end{equation}
\end{small}

\noindent The above formulation holds under any condition with $m_1 \times m_2 > 0$. 
We can observe that the Pearson correlation coefficient, as compared to cosine similarity,
incorporates channel vector centering. This facilitates feature correlation reduction when preserving small numerical
variations, which helps to make the training process more stable.
Such a scale-invariant property can benefit more flexible feature representation.

\section{Complexity Analysis}
In order to highlight the efficiency of our method, we have conducted a comprehensive complexity analysis in Tab.\ref{tab:analyse2}.

\begin{table*}[h]
\caption{Complexity analysis for SOTAs. During once retrieval process, we compute the similarity between a given text and $N$ videos within the gallery. As stated in the main body, we define $N_v$, $N_t$, $N_p$, and $N_r$ as the quantities of frames or segments per video, words per text, patches per frame, and selected videos after reranking, respectively, where  $N_r$ is far less than $N$. }
\vspace{-0.2cm}
\label{tab:analyse2}
\centering
\resizebox{\linewidth}{!}{
\begin{tabular}{@{}lll@{}}
\toprule
Model & Computational Complexity &  Analysis \\ 
\hline
TeachText\cite{teachtext} & $\mathcal{O}(ND)$ & \begin{tabular}[c]{@{}l@{}} When considering either video or text, only one \textbf{\textit{D-dimensional}} \\ feature is required, resulting in a retrieval complexity of $\mathcal{O}(ND)$ \\ for N queries. \end{tabular}    \\ \hline
SEA\cite{sea} & $\mathcal{O}(ND)$ & \begin{tabular}[c]{@{}l@{}} SEA requires 5 different video-text feature pairs in 5 common spaces.\\ Ignore constant item 5, complexity is $\mathcal{O}(ND)$. \end{tabular} \\ \hline
W2VV++\cite{w2vv++}& $\mathcal{O}(ND)$ & \begin{tabular}[c]{@{}l@{}} Only one text/video feature is required, and the  complexity is $\mathcal{O}(ND)$. \end{tabular} \\ \hline
BridgeFormer\cite{bridgeformer} & $\mathcal{O}(ND)$ &  \begin{tabular}[c]{@{}l@{}} Only one text/video feature is required, and the  complexity is $\mathcal{O}(ND)$. \end{tabular} \\ \hline
LAFF\cite{laff} & $\mathcal{O}(ND)$ & \begin{tabular}[c]{@{}l@{}} LAFF requires 8 different video-text feature pairs in 8 common spaces. \\  Ignore constant item 8, complexity is $\mathcal{O}(ND)$. \end{tabular}\\ \hline
CLIP4Clip\cite{clip4clip}& $\mathcal{O}(ND)$ &   \begin{tabular} [c]{@{}l@{}} Only one text/video feature is required, and the  complexity is $\mathcal{O}(ND)$.  \end{tabular}\\ \hline
CenterCLIP\cite{centerclip} & $\mathcal{O}(NN_vD)$ &   \begin{tabular}[c]{@{}l@{}} CenterCLIP divides a video into $N_v$ segments, where each segment \\ is a D-dimension feature. Each text is a \textbf{\textit{D-dimensional}} feature, \\ and the similarity is calculated with all segments, resulting in a \\ complexity of $\mathcal{O}(NN_vD)$. \end{tabular}\\ \hline
TS2-Net\cite{ts2net} & $\mathcal{O}(NN_vD)$ &\begin{tabular}[c]{@{}l@{}} TS2-Net requires  $N_v$ \textbf{\textit{D-dimensional}} features for  frames of a \\ video and one feature for a text, resulting in a complexity of $\mathcal{O}(NN_vD)$.   \end{tabular}   \\ \hline
X-CLIP\cite{xclip} & $\mathcal{O}(N(1+N_vN_t+N_v+N_t)D)$ &\begin{tabular}[c]{@{}l@{}} The text side requires a text feature and $N_t$ word features, and the video \\ side requires a video feature and $N_v$ frame features. Calculate the \\ similarity of text-video, text-frame, word-video, word-frame respectively, \\and the complexity is $\mathcal{O}(N(1+N_vN_t+N_v+N_t)D)$. \end{tabular}   \\ \hline
CLIP-VIP\cite{clipvip}  & $\mathcal{O}(ND)$ &\begin{tabular}[c]{@{}l@{}} Only one text/video feature is required, and the  complexity is $\mathcal{O}(ND)$. \end{tabular}   \\ \hline
X-Pool\cite{xpool}  & $\mathcal{O}(N(N_v+D)D)$ &\begin{tabular}[c]{@{}l@{}}  In the first step, X-Pool requires a \textbf{\textit{D-dimensional}} text feature and $N_v$ \\ \textbf{\textit{D-dimensional}} frame features to calculate similarity, resulting in a \\ complexity of $\mathcal{O}(NN_vD)$. In the second step, X-Pool combines the $Nv$ \\ frame features into a single video feature, which is then passed \\ through a $D \times D$ linear layer, resulting in a complexity of $\mathcal{O}(NDD)$.\\ Overall, the complexity is $\mathcal{O}(N(N_v+D)D)$. \end{tabular}  \\ \hline
EERCF & $\mathcal{O}(ND+N_r(1+N_v+N_p)D)$ &\begin{tabular}[c]{@{}l@{}} Our approach is based on a two-stage retrieval strategy. In the coarse \\ retrieval stage, we calculate the similarities between the \textbf{\textit{D-dimensional}} \\  text feature and all \textbf{\textit{D-dimensional}} text-agnostic video features in the \\ gallery by dot product. The complexity is $\mathcal{O}(ND)$, and $N_r$ top videos \\ are selected for reranking. In the reranking stage,  we utilize TIB module \\ to aggregate $N_v$ \textbf{\textit{D-dimensional}} frame features and $N_p$ \textbf{\textit{D-dimensional}} \\ patch features per video to generate two additional versions of video \\ features, and then recompute the similary between the text and $N_r$ \\ videos, represented by three feature vectors per video. The reranking \\ complexity is $\mathcal{O}(N_r(1+N_v+N_p)D)$.  Overall, the complexity is \\ $\mathcal{O}(ND+N_r(1+N_v+N_p)D)$. \end{tabular}  \\
\bottomrule
\end{tabular}
}
\end{table*}

\subsection{Datasets}

\textbf{MSR-VTT}~\cite{msrvtt} contains 10K videos with 200K captions, which is widely used in video-text retrieval task. The video duration ranges from 10 to 32 seconds, and corresponding texts vary a lot in length and content, which supports our motivation for learning adaptive and compact video features for retrieval. Following the popular settings, We adopt two types of dataset partitions. One is the official data split, namely MSRVTT-3K-Test, dividing the dataset into 6,513 videos for training, 497 videos for validation, and 2,990 videos for test. The other is adopted by recent methods~\cite{clip4clip,xpool}, which use roughly 9K videos for training and the remaining 1K videos for test, called MSRVTT-1K-Test.

\textbf{VATEX}~\cite{vatex} includes 34,991 videos with multilingual annotations. Following HGR~\cite{hrg}, we use 25,991 videos as the training set, 1,500 videos as the validation set, and 1,500 videos as the test set, respectively.

\textbf{MSVD}~\cite{msvd} is composed of 1,970 videos with 40 captions per video.  Following the official setting, we use 1,200, 100, and 670 videos for training, validating, and testing.

\textbf{ActivityNet}~\cite{activitynet} contains 20K YouTube videos with 100K sentences. Following~\cite{multimodal}, we concatenate all the captions of a video to form a paragraph and evaluate our model for video-paragraph retrieval on ‘val1’ split.

\subsection{More Results on Benchmarks}
We a series of detailed experimental data for exploring the hyperparameters  top-k in Tab.\ref{tab:topk}.
We provide complete  experimental results on MSRVTT-3K-Test\cite{msrvtt}, VATEX\cite{vatex}, MSVD\cite{msvd} and ActivityNet\cite{activitynet}.

\begin{table}[!h]
\caption{Ablation study of the top-k reranking parameter in EERCF on MSRVTT-1k-Test dataset.}
\vspace{-0.2cm}
\label{tab:topk}
\centering
\resizebox{\linewidth}{!}{
\begin{tabular}{@{}lllllllll@{}}
\toprule
\multirow{2}{*}{top-k} & \multicolumn{4}{c}{$t2v$ Retrieval} & \multicolumn{4}{c}{$v2t$ Retrieval} \\ \cline{2-9} 
 & R@1 & R@5 & R@10 & Mean & R@1 & R@5 & R@10 & Mean \\ \hline
10 & 47.7 & 71.4 & 78.5 & 65.9 & 45.3 & 71.8 & 80.6 & 65.9 \\
20 & 48.0 & 72.4 & 81.7 & 67.4 & 45.3 & 72.8 & 82.7 & 66.9 \\
50 & 47.8 & 74.1 & 84.1 & 68.7 & 44.7 & 74.2 & 83.9 & 67.6 \\
100 & 47.8 & 73.2 & 82.7 & 67.9 & 45.4 & 73.1 & 83.7 & 67.4 \\ 
\bottomrule
\end{tabular}
}
\end{table}


\begin{table}[H]
\caption{Performance of SOTAs on MSRVTT-3K-Test.}
\vspace{-0.2cm}
\label{tab:msrvttfull_supple}
\centering
\resizebox{\linewidth}{!}{
\begin{tabular}{@{}lllllllll@{}}
\hline
\multicolumn{1}{l}{\multirow{2}{*}{Model}} & \multicolumn{4}{c}{$t2v$ Retrieval} & \multicolumn{4}{c}{$v2t$ Retrieval} \\ \cline{2-9} 
\multicolumn{1}{c}{} & R@1 & R@5 & R@10 & Mean & R@1 & R@5 & R@10 & Mean \\ \hline
TeachText\cite{teachtext} & 15.0 & 38.5 & 51.7 & 35.1 & 25.3 & 55.6 & 68.6 & 49.8 \\
SEA\cite{sea} & 19.9 & 44.3 & 56.5 & 40.2 & - & - & - & - \\
W2VV++\cite{w2vv++} & 23.0 & 49.0 & 60.7 & 44.2 & - & - & - & - \\
LAFF\cite{laff} & 29.1 & 54.9 & 65.8 & 49.9 & - & - & - & - \\ 
CLIP4Clip$\dagger$\cite{clip4clip} & 29.4 & 54.9 & 65.8 & 50.0 & 50.8 & 78.5 & 87.0 & 72.1 \\
TS2-Net$\dagger$\cite{ts2net} & 29.9 & 56.4 & 67.3 & 51.2 & 53.0 & 82.5 & 89.8 & 75.1 \\
X-CLIP$\dagger$\cite{xclip} & 31.2  & 57.4  &  68.1  &  52.2  & 56.1  & 84.3  &  92.3  & 77.6  \\
 \hline
EERCF & 31.5  & 57.4 & 67.6  & 52.2  & 56.7  & 84.7  & 92.2  & 77.9  \\ \hline
\end{tabular}
}
\end{table}

\begin{table}[H]
\caption{Performance of SOTAs on VATEX.}
\vspace{-0.2cm}
\label{tab:vatex_supple}
\centering
\resizebox{\linewidth}{!}{
\begin{tabular}{@{}lllllllll@{}}
\hline
\multicolumn{1}{l}{\multirow{2}{*}{Model}} & \multicolumn{4}{c}{$t2v$ Retrieval} & \multicolumn{4}{c}{$v2t$ Retrieval} \\ \cline{2-9} 
\multicolumn{1}{c}{} & R@1 & R@5 & R@10 & Mean & R@1 & R@5 & R@10 & Mean \\ \hline
TeachText\cite{teachtext} & 53.2 & 87.4 & 93.3 & 78.0 & 64.7 & 91.5 & 96.2 & 84.1 \\
SEA\cite{sea} & 52.4 & 90.2 & 95.9 & 79.5 & - & - & - & -\\
W2VV++\cite{w2vv++} & 55.8 & 91.2 & 96.0 & 81.0 & - & - & - & -\\
LAFF\cite{laff} & 59.1 &  91.7 &  96.3   & 82.4 & - & - & - & -\\ 
X-CLIP$\dagger$\cite{xclip} & 62.1 & 90.8 & 95.3 & 82.7 & 78.1 & 98.0 & 99.2 & 91.8 \\
CLIP4Clip$\dagger$\cite{clip4clip} & 61.6 & 91.1 & 95.8  & 82.8 & 78.2 & 97.3 & 99.2 & 91.6 \\
TS2Net$\dagger$\cite{ts2net} & 61.1 & 91.5 & 96.0 & 82.9 &  78.7  & 97.8 & 99.3 &  91.9  \\ 
DRL\cite{wang2022disentangled} 63.5 & 91.7 & 96.5 & 83.9 &  77.0  & 98.0 & 99.4 &  91.5 \\
\hline
EERCF &  62.6  & 91.5 & 95.8  &  83.3  & 77.5 &  98.3  & 99.5 & 91.8 \\ 
\hline
\end{tabular}
}
\end{table}

%

\begin{table}[H]
\caption{Performance of SOTAs on MSVD.
\vspace{-0.2cm}}
\label{tab:msvd_supple}
\centering
\resizebox{\linewidth}{!}{
\begin{tabular}{@{}lllllllll@{}}
\hline
\multirow{2}{*}{Model} & \multicolumn{4}{c}{$t2v$ Retrieval} & \multicolumn{4}{c}{$v2t$ Retrieval} \\ \cline{2-9} 
 & R@1 & R@5 & R@10 & Mean & R@1 & R@5 & R@10 & Mean \\ \hline
TeachText\cite{teachtext} & 25.4 & 56.9 & 71.3 & 51.2 & 27.1 & 55.3 & 67.1 & 49.8 \\
SEA\cite{sea} & 34.5 & 68.8 & 80.5 & 61.3 & - & - & - & - \\
W2VV++\cite{w2vv++} & 37.8 & 71.0 & 81.6 & 63.5 & - & - & - & - \\
LAFF\cite{laff} & 45.4 & 70.6 & 84.6 & 66.9 & - & - & - & -\\
TS2-Net$\dagger$\cite{ts2net} & 44.6 & 75.8 & 84.5 & 68.3 & 55.6 & 77.2 & 83.0 & 71.9 \\
CLIP4Clip$\dagger$\cite{clip4clip} & 45.6 & 76.1 & 84.3 & 68.9 & 63.3  & 85.4  & 92.6  & 80.1  \\
X-CLIP$\dagger$\cite{xclip} & 46.7 & 76.8 & 85.5  & 69.7 & 59.0 & 83.1 & 86.5 & 76.2 \\
X-Pool\cite{xpool} & 47.2  & 77.4  & 86.0  &  70.2  & - & - & - & - \\ 
DRL\cite{wang2022disentangled} & 48.3  & 79.1  & 87.3  &  71.6   & 62.3 & 86.3 & 92.2 & 80.3 \\
\hline

EERCF & 47.0  &  77.5  & 85.4 & 70.0  & 62.0  &  87.1  & 94.3 & 81.1  \\ \hline
\end{tabular}
}
\end{table}


%
%
\begin{table}[H]
\caption{Performance of SOTAs on ActivityNet.}
\vspace{-0.2cm}
\label{tab:activitynet_supple}
\centering
\resizebox{\linewidth}{!}{
\begin{tabular}{@{}lllllllll@{}}
\hline
\multirow{2}{*}{Model} & \multicolumn{4}{c}{$t2v$ Retrieval} & \multicolumn{4}{c}{$v2t$ Retrieval} \\ \cline{2-9} 
 & R@1 & R@5 & R@10 & Mean & R@1 & R@5 & R@10 & Mean \\ \hline
TeachText\cite{teachtext} & 23.5 & 57.2 & 96.1 & 58.9 & 23.0 & 56.1 & 95.8 & 58.3 \\
TS2Net$\dagger$\cite{ts2net} & 37.3 & 69.9 & 83.2 & 63.5 & 41.3 & 72.1 & 85.1 & 66.2 \\
CLIP4Clip$\dagger$\cite{clip4clip} & 39.7 & 71.0 & 83.4 & 64.7 & 39.5 & 71.4 & 83.6 & 64.8 \\
CenterCLIP\cite{centerclip} & 43.5  &  75.1  & 85.4  &  68.0  & 44.1  & 75.1  & 86.0 & 68.4 \\
X-CLIP$\dagger$\cite{xclip} &  44.4  &  74.6  & 85.1 &  68.0  &  44.4  & 74.7 & 86.3  & 68.5  \\
DRL\cite{wang2022disentangled} & 44.2 & 74.5 & 86.1 & 68.3 & 42.2 & 74.0 & 86.2 & 67.5 \\
\hline
EERCF & 43.1 & 74.5 &  86.0   & 67.9  & 42.9 &  76.2  & 87.7  &  68.9  \\ \hline
\end{tabular}
}
\end{table}


%
%

\subsection{More Visual Cases}
\begin{figure*}[!th]
\centering
\includegraphics[width=0.77\linewidth]{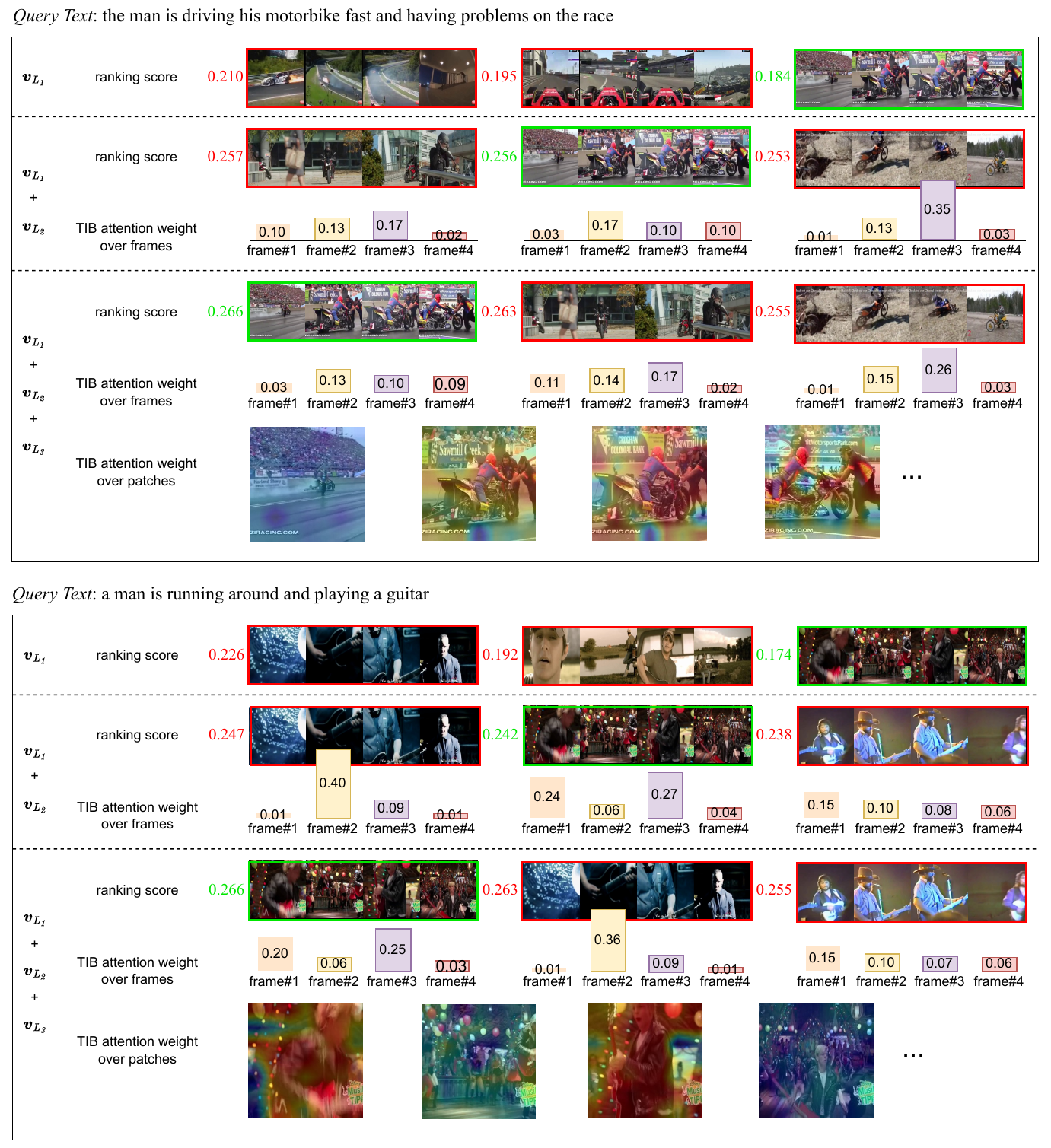}
\caption{\textbf{Visualization of good cases on MSRVTT-1K-Test.} Green boxes mean the
groundtruth video corresponding to the query text, and red
boxes denote confused videos.}
\label{fig:vis_posi}
\end{figure*}

\begin{figure*}[!th]
\centering
\includegraphics[width=0.77\linewidth]{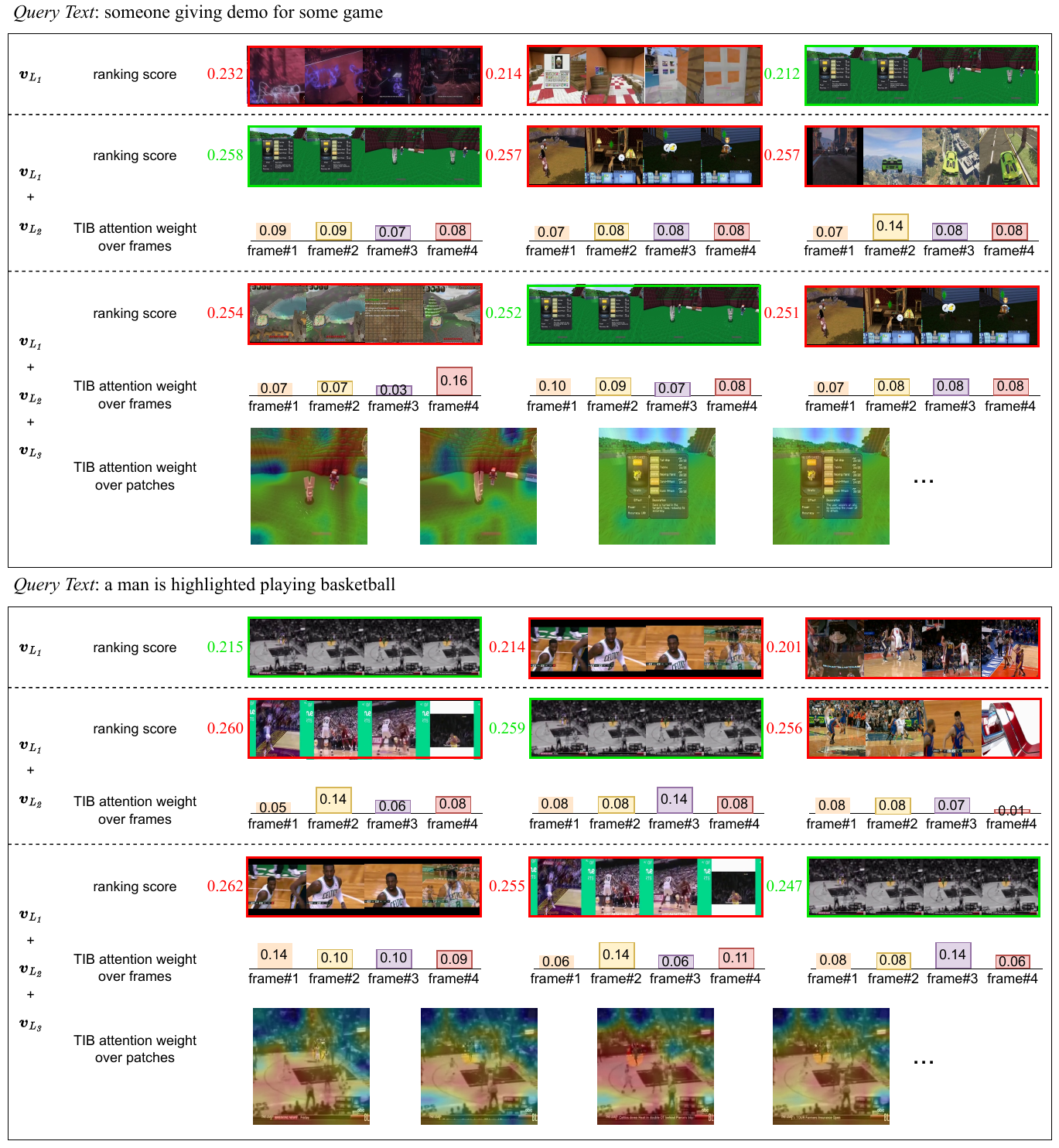}
\caption{\textbf{Visualization of bad cases on MSRVTT-1K-Test.}  
In the first case, the query ``someone giving demo for some game" is used to describe overall scene rather than details, making it arduous to capture the similarity between the query and the patches. Consequently, we can see the patches' heatmap is scattered, and we guess this leads to a decline in the search ranking. 
In the second case, we point out some noise is introduced by the annotators in the video-text pairs. The query term "highlighted" can be interpreted as a close-up  "meaningful instant" shot. We can see the videos ranked higher are actually  relevant to the query at the frame or patch level, which can prove that EERCF has good robustness against noise. }
\label{fig:vis_nega}
\end{figure*}

We provide additional cases to facilitate better comprehension of our CLIP-TIP model in Fig.\ref{fig:vis_posi} and Fig.\ref{fig:vis_nega}. In particular, there are some exemplars showcasing model failure.

\bibliography{main}

\end{document}